%% file: acl2019.tex
\documentclass[11pt,a4paper]{article}
\usepackage[hyperref]{acl2019}
\usepackage{times}
\usepackage{latexsym}
\usepackage{booktabs}
\usepackage{amsmath}

\usepackage{url}
\usepackage{graphicx}

\usepackage{latexsym}
\usepackage{graphicx}
\usepackage{makecell}

\aclfinalcopy 

\title{
Dr.Quad at MEDIQA 2019: Towards Textual Inference \\and Question Entailment using contextualized representations
}

\author{Vinayshekhar Bannihatti Kumar \thanks{\quad equal contribution}\hspace{2mm}
    Ashwin Srinivasan\footnotemark[1] \hspace{2mm}
  Aditi Chaudhary\footnotemark[1]  \\
   {\bf James Route}  \hspace{2mm}
   {\bf Teruko Mitamura} \hspace{2mm}
  {\bf Eric Nyberg} \\
  $\{vbkumar, ashwinsr, aschaudh, jroute,teruko, ehn \}$@cs.cmu.edu \\
  Language Technologies Institute \\
  Carnegie Mellon University
 }

\date{}

\begin{document}
\maketitle
\begin{abstract}
This paper presents the submissions by Team Dr.Quad to the ACL-BioNLP 2019 shared task on Textual Inference and Question Entailment in the Medical Domain. Our system is based on the prior work \citet{liu2019multi} which uses a multi-task objective function for textual entailment. In this work, we explore different strategies for generalizing state-of-the-art language understanding models to the specialized medical domain. Our results on the shared task demonstrate that incorporating domain knowledge through data augmentation is a powerful strategy for addressing challenges posed by specialized domains such as medicine.  
\end{abstract}

\input{Introduction.tex}

\input{Experiments.tex}
\input{Conclusion.tex}
\bibliographystyle{acl_natbib}
\bibliography{acl2019}
\end{document}

%% file: Introduction.tex
\section{Introduction}
\label{intro}
 The ACL-BioNLP 2019 \cite{aclmediqa} shared task focuses on improving the following three tasks for medical domain: 1) Natural Language Inference (NLI) 2) Recognizing Question Entailment (RQE) and 3) Question-Answering re-ranking system. Our team has made submissions to all the three tasks. We note that in this work we focus more on the task 1 and task 2 as improvements in these two tasks reflect directly on the task 3. However, as per the shared task guidelines, we do submit one model for the task 3 to complete our submission. 

Our approach for both task 1 and task 2 is based on the state-of-the-art natural language understanding model MT-DNN \cite{liu2019multi}, which combines the strength of multi-task learning (MTL) and language model pre-training. MTL in deep networks has shown performance gains when related tasks are trained together resulting in better generalization to new domains \cite{ruder2017overview}. Recent works such as BERT \cite{devlin2018bert}, ELMO \cite{peters2018deep} have shown the efficacy of learning universal language representations in providing a decent warm start to a task-specific model, by leveraging large amounts of unlabeled data. MT-DNN uses BERT as the encoder and uses MTL to fine-tune the multiple task-specific layers. This model has obtained state-of-the-art results on several natural language understanding tasks such as SNLI \cite{bowman2015large}, SciTail \cite{khot2018scitail} and hence forms the basis of our approach. For the task 3, we use a simple model to combine the task 1 and task 2 models as shown in \S \ref{qa}. 

As discussed above, state-of-the-art models using deep neural networks have shown significant performance gains across various natural language processing (NLP) tasks. However, their generalization to specialized domains such as the medical domain still remains a challenge.  \citet{romanov2018lessons} introduce a new dataset MedNLI, a natural language inference dataset for the medical domain and show the importance of incorporating domain-specific resources. Inspired by their observations, we explore several techniques of augmenting domain-specific features with the state-of-the-art methods. We hope that the deep neural networks will help the model learn about the task itself and the domain-specific features will assist the model in tacking the issues associated with such specialized domains. For instance, the medical domain has a distinct sublanguage \cite{friedman2002two} and it presents challenges such as abbreviations, inconsistent spellings, relationship between drugs, diseases, symptoms. 

Our resulting models perform fairly on the unseen test data of the ACL-MediQA shared task. On Task 1, our best model achieves +14.1 gain above the baseline. On Task 2, our five-model ensemble achieved +12.6 gain over the baseline and for Task 3 our model achieves a a +4.9 gain.

%% file: Experiments.tex
\section{Approach}
\label{appraoch}
In this section, we first present our base model MT-DNN \cite{liu2019multi} which we use for both Task 1 and Task 2 followed by a discussion on the different approaches taken for natural language inference (NLI) (\S \ref{nli}), recognizing question entailment (RQE) (\S \ref{rqe}) and question answer (QA) (\S \ref{qa}). 

\begin{figure}[ht]
\includegraphics[width=0.5\textwidth, height=60mm]{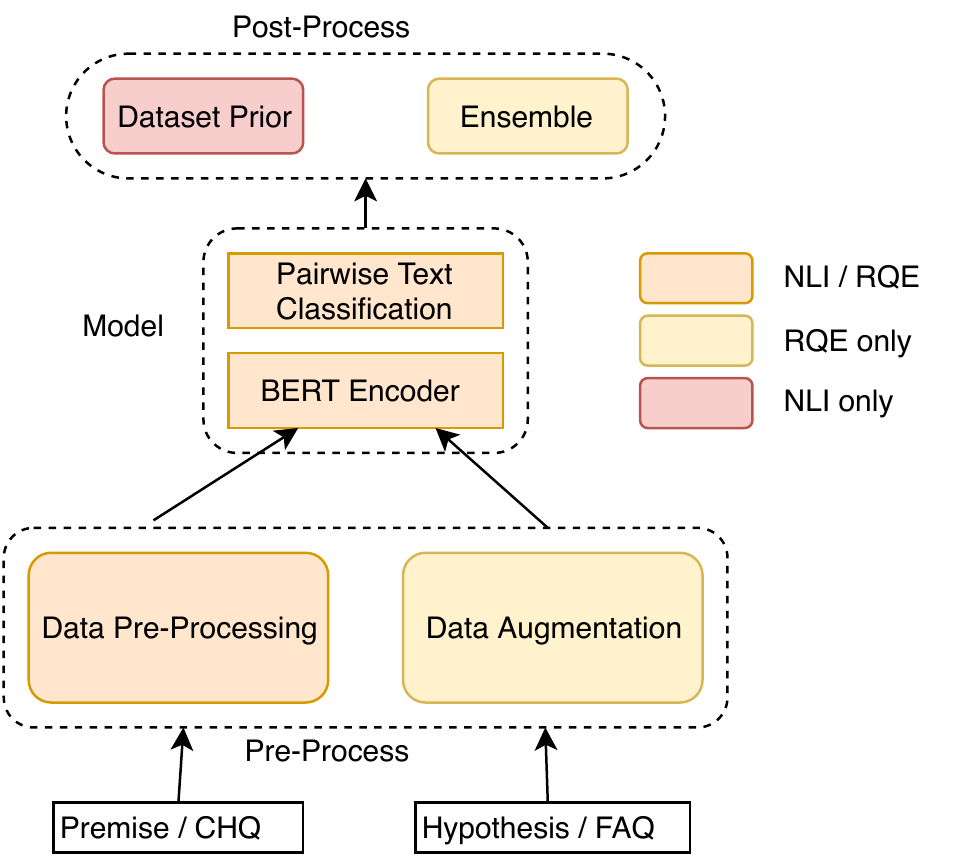}
\caption{\label{fig:system-architeture} System overview for NLI and RQE task}.
\end{figure}
\subsection{Task 1 and Task 2 Formulation}

Formally, we define the problem of textual entailment as a multi-class classification task. Given two sentences \textbf{a} =\textit{$a_1, a_2 ..., a_n$}  and \textbf{b} = \textit{$b_1, b_2, ..., b_m$}, the task is to predict the correct label. For NLI, \textbf{a} refers to the \textit{Premise} and \textbf{b} refers to the \textit{Hypothesis} and the label set comprises of \textit{entailment, neutral,  contradiction}. For RQE, \textbf{a} refers to the \textit{CHQ} and \textbf{b} refers to the \textit{FAQ} and the label set comprises of \textit{True, False}.

\subsection{ Model Architecture}
A brief depiction of our system is shown in Figure~\ref{fig:system-architeture}. We represent components which were used for both NLI and RQE in Orange. An example of this is the Data Pre-processing component. The RQE only components are shown in yellow (eg. Data Augmentation). The components which were used only for the NLI modules are shown in Pink (eg. Dataset Prior). We base our model on the state-of-the-art natural language understanding model MT-DNN \cite{liu2019multi}. MT-DNN is a hierarchical neural network model which combines the advantages of both multi-task learning and pre-trained language models. Below we describe the different components in detail.
\paragraph{Encoder:} Following BERT \cite{devlin2018bert}, each sentence pair is separated by a [SEP] token. It is then passed through a lexicon encoder which represents each token as a continuous representation of the word, segment and positional embeddings. A multi-layer bi-directional transformer encoder \cite{vaswani2017attention} transforms the input token representations into the contextual embedding vectors. This encoder is then shared across multiple tasks.
\paragraph{Decoder:} We use the \textit{Pairwise text classification output} layer \cite{liu2019multi} as our decoder. Given a sentence pair (\textbf{a},\textbf{b}), the above encoder first encodes them into \textbf{u} and \textbf{v} respectively. Then a K-step reasoning is performed on these representations to predict the final label. The initial state is given by $\mathbf{s} = \sum_j \alpha_j \mathbf{u_j}$ where $\alpha_j = \frac{\exp(\mathbf{w}^T \mathbf{u_j})}{\sum_i \exp(\mathbf{w_1}^T \mathbf{u_i})}$. On subsequent iterations $k \in [1,K-1]$, the state is $\mathbf{s}^k = GRU(\mathbf{s}^{k-1}, \mathbf{x}^k)$ where $\mathbf{x}_k = \sum_j \beta_j \mathbf{v_j} $ and $\beta_j = softmax(\mathbf{s}_{k-1}\mathbf{w_2}^T\mathbf{v})$. Then a single-layer classifier predicts the label at each iteration $k$:
\[P^k = softmax(\mathbf{w_3}^T[\mathbf{s}^k;\mathbf{x}^k;|\mathbf{s}^k-\mathbf{x}^k|;\mathbf{s^k}.\mathbf{x^k}])\] Finally, all the scores across the $K$ iterations are averaged for the final prediction. We now describe the modifications made to this model for each respective task.
\subsection{Natural Language Inference}
\label{nli}
This task consists of identifying three inference relations between two sentences: Entailment, Neutral and Contradiction
\paragraph{Data:} The data is based off the MedNLI dataset introduced by \citet{romanov2018lessons}. The statistics of the dataset can be seen in Table \ref{table:MedNLI}.
\begin{table}[t!]
\begin{center}\resizebox{0.4\textwidth}{!}{
\begin{tabular}{|l|l|l|l|}
   & \bf Train & \bf Validation & \bf Test \\ 
  \toprule
Entailment & 3744 & 465 &  474 \\
Contradiction & 3744 & 465 & 474 \\
Neutral &  3744 & 465 & 474 \\
\bottomrule
\end{tabular}
}
\end{center}
\caption{\label{table:MedNLI} The number of train and test instances in each of the categories of the NLI dataset.}
\end{table}

\paragraph{Data Pre-Processing:} On manual inspection of the data, we observe the presence of abbreviations in the premise and hypothesis. Since lexical overlap is a strong indicator of entailment by virtue of pre-trained embeddings on large corpora, the presence of abbreviations makes it challenging. Therefore, we expand the abbreviations using the following two strategies:
\begin{enumerate}
    \item \textit{Local Context:} We observe that often an abbreviation is composed of the first letters of contiguous words.  Therefore, we first construct potential abbreviations by concatenating first letter of all words in an sequence, after tokenization. For instance, for the premise shown below we get \{CXR, CXRS, XRS, CXRSI, XRSI, RSI, etc\}.  This is done for both the premise and the hypothesis. We then check if this n-gram exists in the hypothesis (or the premise). If yes, then we replace that abbreviation with all the words that make up the n-gram. Now the model has more scope of matching two strings lexically. We demonstrate an example below:
    
    \textbf{Premise: }Her \textbf{CXR} was clear and it did not appear she had an infection.\\
     \textbf{Hypothesis: } \textbf{Chest X-Ray} showed infiltrates.
     
    \textbf{Premise Modified: } Her \textbf{Chest X-Ray} was clear and it did not appear she had an infection.
    
    \item \textit{Gazetteer:} If either the premise/hypothesis does not contain the abbreviation expansion or contains only partial expansion, the \textit{Local Context} technique will fail to expand those abbreviations. Hence, we use an external gazetteer extracted from CAMC\footnote{https://www.camc.org/} to expand commonly occurring medical terms. There were 1373 entries in the gazetteer, covering common medical and clinical expansions. For instance,\\
    \textbf{Premise: }On arrival to the \textbf{MICU} , patient is hemodynamically stable .\\
    \textbf{Premise Modified: } On arrival to the \textbf{Medical Intensive Care Unit} , patient is hemodynamically stable .
\end{enumerate}

We first performed the local context replacement as they are more specific to a given premise-hypothesis pair. If there was no local context match, then we did a gazetteer lookup. It is to be noted that one abbreviation can have multiple expansions in the gazatteer and thus we hypothesized that local context should get preference while expanding the abbreviation.

\paragraph{Training Procedure:} For training the MT-DNN model,  we use the same hyper-parameters provided by the authors \cite{liu2019multi}. We train model for 4 epochs and early stop when the model reaches the highest validation accuracy.

\paragraph{Baselines:} We use the following baselines similar to \citet{romanov2018lessons}.
\begin{itemize}
\item \textit{CBOW:} We use a Continuous-Bag-Of-Words (CBOW) model as our first baseline. We take both the premise and the hypothesis and sum the word embeddings of the respective statements to form the input layer to our CBOW model. We used 2 hidden layers and used softmax as the decision layer.

\item \textit{Infersent:}   Inferesent is a sentence encoding model which encodes a sentence by doing a max-pool on all the hidden states of the LSTM across time steps. We follow the authors of \citet{romanov2018lessons} by using shared weights LSTM cell to get the sentence representation of the premise(U) and the hypothesis(V). We feed these representations U and V to an MLP to perform a 3 way prediction. For our experiments, we use the pre-trained embeddings trained on the MIMIC dataset by \citet{romanov2018lessons}. We used the same hyperparameters.

\item \textit{BERT:} Since MT-DNN is based off of the BERT \cite{devlin2018bert} model as the encoder, we also compare results using just the pre-trained BERT. We used \textit{bert-base-uncased} model which was trained for 3 epochs with a learning rate of 2e-5 and a batch size of 16 with a maximum sequence length of 128. WE used the last 12 pre-trained layers of the model.
\end{itemize}



\subsubsection{Results and Discussion}
\label{results:nli}
In this section we discuss the results of all of our experiments on the NLI task. 

\paragraph{Ablation Study:} First, we conduct an ablation study to study the effect of abbreviation expansion. Table ~\ref{table:abbreviation} shows the results of the two abbreviation expansion techniques for the Infersent model. We observe the best performance with the \textit{Gazetteer} strategy. This is because most sentences in the dataset did not have the abbreviation matched through the local context match.  Since expanding abbreviations helped increase lexical overlap, going forward we use the expanded abbreviation data for all our experiments henceforth.
\begin{table}[t!]
\small
\begin{center}
\begin{tabular}{|l|l|l|l|}
 \bf Model Ablation & \bf Accuracy \\ \toprule
 Infersent & 78.8 +/- 0.06 \\
 Infersent + Local-Context  &  78.8 +/- 0.02  \\
 Infersent + Local-Context + Gazetteer & 78.5 +/- 0.36 \\
 Infersent + Gazetteer & \textbf{79.1 +/ 0.14} \\
\bottomrule
\end{tabular}
\end{center}
\caption{\label{table:abbreviation} The results reported in the table is mean and variance of the models averaged on 3 runs using different random seeds.}
\end{table}
Table \ref{table:confusion-matrix-infersent} shows the confusion matrix for the Infersent model. The rows represent the ground truth and the columns represent the predictions made by us.  We can see that the model is most confused about the \textit{entailment} and \textit{neutral} classes. 82 times the model predicts \textit{neutral} for \textit{entailment} and 85 times vice versa. In order to address this issue, we add a prior on the dataset as a post processing step.

\begin{table}[h]
\small
\begin{center}
\begin{tabular}{|l|l|l|l|l|}
 & \bf Contradiction & \bf Entailment & \bf Neutral\\ \toprule
 Contradiction & 396 & 43 & 26 \\
 Entailment & 30 & 353 & \textbf{82} \\
 Neutral & 23 & \textbf{85} & 357 \\
\bottomrule
\end{tabular}
\end{center}
\caption{\label{table:confusion-matrix-infersent} Confusion matrix  for NLI classes for Infersent model. Rows denote the true labels and columns denote the model predictions.}
\end{table}

\paragraph{Prior on the dataset:}
Our dataset analysis on the validation set revealed that there were three hypothesis for a given premise with mutually exclusive labels. Since we know that for a given premise there can only be one entailment because of the nature of the dataset, we post-process the model predictions to add this constraint. For each premise we collect the prediction probability for each of the hypothesis and pick the hypothesis having the highest probability for entailment. We perform the same selectional preference procedure on the remaining two classes. Such a post-processing ensures that each premise always has three hypotheses with mutually exclusive labels.

Table \ref{tab:nliresults} documents the results of the different models on the validation set. We observe that our method gives the best performance among the three baselines. Based on these results, our final submission on the unseen data can be seen in the last row.
\begin{table}[h]
\small
\begin{center}
\begin{tabular}{|l|l|l|l|}
 \bf Model Ablation & \bf Accuracy \\ \toprule
 CBOW &  74.7   \\
 Infersent & 79.1 \\
 BERT & 80.4 \\
 Ours & \textbf{82.1} \\
 \midrule
 \cite{aclmediqa} (Unseen Test) & 71.4 \\
  Ours (Unseen Test) &  79.6\\
   Ours (Unseen Test) + Prior &  \textbf{85.5}\\
\bottomrule
\end{tabular}
\end{center}
\caption{\label{tab:nliresults} NLI results on the validation set.}
\end{table}

\subsubsection{Error Analysis}
\begin{table*}[ht]
\begin{center}\resizebox{\textwidth}{!}{
  \begin{tabular}{|l|l|l|l|}

 \textbf{Type} & \textbf{CHQ} & \textbf{FAQ} & \textbf{Label}\\
  \toprule
 Train & What is the treatment for  & What is the treatment for & True\\
  & tri-iodothyronine thyrotoxicosis?&  T3 (triiodothyronine) thyrotoxicosis? &  \\
  & Do Coumadin and Augmentin interact? & How do you inject the bicipital tendon? & False \\
  \midrule
  Validation & sepsis. Can sepsis be prevented. & Who gets sepsis? & True \\
  & Can someone get this from a hospital? &  &  \\
  & medicine and allied. I LIKE TO KNOW & What is an Arrhythmia? & False \\
  & RECENT THERAPY ON 
          ARRHYTHMIA OF HEART &  &  \\
    \bottomrule
  \end{tabular}
  }
  \caption{Examples of question entailment from train and validation set.}
  \label{tab:datarqe}
  \end{center}
\end{table*}

We perform qualitative analysis of  our model and bucket the errors into the following categories.
\input{error-analysis.tex}
\begin{enumerate}
    \item \textbf{Lexical Overlap:} From Table \ref{table:qualitative}, we see that there is a high lexical overlap between the premise and hypothesis, prompting our model to falsely predict \textit{entailment}.
    \item \textbf{Disease-Symptom relation:} In the second example, we can see that our model lacks sufficient domain knowledge to relate \emph{hyperglycemia} (a symptom) to \emph{diabetes} (a disease). The model interprets these to be two unrelated entities and labels as \textit{neutral}.
    \item \textbf{Drug-disease relation:} In the final example we see that our model doesn't detect that the drug names in the premise actually entail the condition in hypothesis. 
\end{enumerate}

These examples show that NLI in the medical domain is very challenging and requires integration of domain knowledge with respect to understanding complex drug-disease or symptom-disease relations.
\subsection{Recognizing Question Entailment}

\label{rqe}
This task focuses on identifying entailment  between  two  questions and is referred as recognizing question entailment (RQE). The task is defined as : "a question A entails a question B if every answer to B is also a complete or partial answer to A".  One of the questions is called CHQ and the other FAQ.

\paragraph{Data:} The data is based on the RQE dataset collected by \citet{abacha2016recognizing}. The dataset statistics can be seen in Table \ref{table:RQE}.
\begin{table}[h]
\begin{center}
\begin{tabular}{|l|l|l|}
 \textbf{Label}& \bf Train Set & \bf Validation set \\ \toprule
True & 4655 & 129 \\
False & 3933 & 173 \\
\bottomrule
\end{tabular}
\end{center}
\caption{\label{table:RQE} The number of train and validation instances in each of the categories of the RQE dataset.}
\end{table}

\paragraph{Pre-Processing:} Similar to the NLI task, we pre-process the data to expand any abbreviations in the CHQ and FAQ.

\paragraph{Training Procedure:}
The multi-task MT-DNN model gave the best performance for the NLI task, which motivated us to use it for the RQE task as well. We use the same hyperparamters as \citet{liu2019multi} and train the model for 3 epochs.

\paragraph{Baselines:} We compare our model with the following baselines:
\begin{itemize}
    \item \textit{SVM:}  Similar to \citet{abacha2016recognizing}, we use a feature based model SVM and Logistic Regression for the task of question entailment.  We extract the features presented in \citet{abacha2016recognizing} to the best of our abilities. Their model uses lexical features such as word overlap, bigram proportion,  Named Entity Recognition (NER) features and features from the Unified Medical Concepts (UMLS) repository. Due to access issues, we only use the i2b2 \footnote{https://www.i2b2.org/NLP/DataSets/} corpus for extracting the NER features. 
    
    \item \textit{BERT:} Like before, we compare our model with the pre-trained BERT model. For this task, we used the \textit{bert-base-uncased} model and fine-tuned the last 12 layers for 4 epochs with learning rate 2e-5. A batch size of 16 was used.
\end{itemize}

\subsubsection{Distribution Mismatch Challenges }
\label{rqedomain}
The RQE dataset posed many unique challenges, the main challenge being that of distribution mismatch between the train and validation distribution.  Table \ref{tab:datarqe} shows some examples from the training and validation set which illustrate these challenges. We observe that in the training set, entailing examples always have high lexical overlap. There were about 1543 datapoints in the training set where the CHQ and FAQ were exact duplicates. The non-entailing examples in the training set are completely un-related and hence the negative examples are not strong examples. Whereas in the validation set the negative examples also have lexical overlap. Furthermore, the nature of text in the validation set is more informal with inconsistent casing, punctuation and spellings whereas the training set is more structured. Furthermore, the length of the CHQ in the validation set is much longer than those observed in the training set. Therefore, we design our experimental settings based on these observations. 

\begin{figure*}
 \centering
  \includegraphics[width=\textwidth]{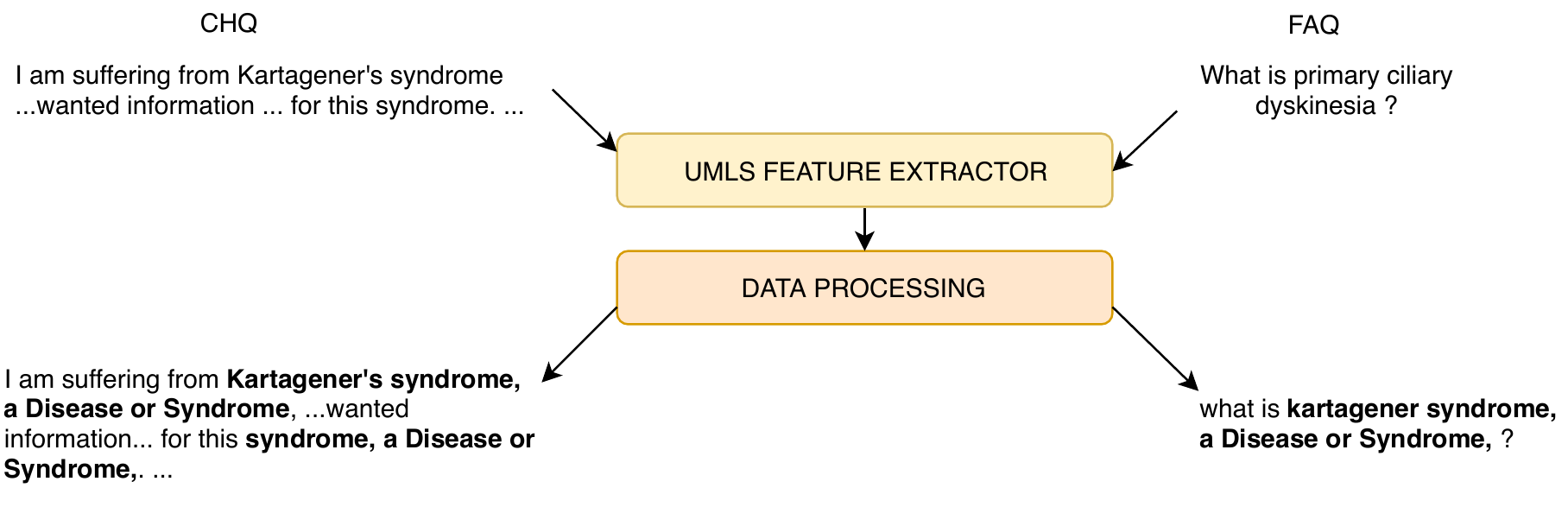}
  \caption{Data augmentation using domain knowledge for RQE.}
  \label{fig:rqedata}
\end{figure*}


\subsubsection{Data Augmentation}
In order to address these challenges, we attempt to create synthetic data which is similar to our validation set. Another motivation for data augmentation was to increase the training size because neural networks are data hungry.
Since most deep neural models rely on lexical overlap as strong indicator of entailment, we therefore use the UMLS features to augment our training set, but such that they help disambiguate the false positives. We use the following procedure for data augmentation:

\begin{enumerate}
\item We retrieve UMLS features for each question in the training, validation and test datasets, using the MetaMap \footnote{https://metamap.nlm.nih.gov} classifier. 
\item We use the retrieved concept types and canonical names to create a new question-pair with the same label as shown in Figure \ref{fig:rqedata}, where the phrase \textit{primary ciliary dyskinesia} has been replaced by its canonical name \textit{kartaganer syndrome} and concept type \textit{Disease or Syndrome}. Since BERT and MT-DNN have been trained on vast amount of English data including Wikipedia, the models are sensitive to language structure. Therefore, while augmenting data with UMLS features, we attempt to maintain the language structure, as demonstrated in Figure \ref{fig:rqedata}. Since UMLS provides the canonical features for each phrase in the sentence, we replace the found phrase with the following template \textit{$<$ UMLS Canonical name $>$, a $<$UMLS Concept Type$>$}. 
\end{enumerate}

Along with the synthetic data, we also experiment with another question entailment dataset Quora-Question Pairs (QQP). We describe the different training data used in our experiments: 

\begin{enumerate}
    \item \textit{Orig:} Using only the provided training data.
    \item \textit{DataAug:} Using the validation set augmented with the UMLS features as discussed above. The provided training data was not used in this setting because of distribution mismatch. Despite the validation set being low-resources (300 sentences), MT-DNN has shown the capability of domain adaptation even in low-resource settings.
    \item \textit{ QQP:} Quora Question pair  \footnote{https://data.quora.com/First-Quora-Dataset-Release-Question-Pairs}(QQP) is a dataset which was released to identify duplicate questions on Quora. Questions are considered duplicates if the answer to one question can be be used as the answer to another question. We hypothesized that jointly training the model with the Quora-Question Pairs dataset should help as it is closest to our RQE dataset in terms of online forum data. We choose a subset of approx. 9k data points from QQP as this dataset has 400k training data points, in order to match the data points from the RQE training data. Along with this we use the validation set to train our model.
    \item \textit{Paraphrase:} Generated paraphrases of the \textit{DataAug} using an off-the-shelf tool \footnote {https://paraphrasing-tool.com}. This was inspired by the observation that validation set was in-domain but since it was low-resourced, this tool provides a cheap way of creating additional  artificial dataset.
\end{enumerate}

\begin{table*}[htb]
\begin{center}\resizebox{\textwidth}{!}{
\begin{tabular}{|l|l|l|}

Lexical Overlap & \makecell[l]{\textbf{CHQ} \\   \textbf{FAQ}\\ \textbf{Ground truth} \\ \textbf{Prediction}} & \makecell[l]{Please i want to know the cure to Adenomyosis... I want to see a specialist doctor to help me out. \\ Do I need to see a doctor for Adenomyosis ? \\ False \\ True }  \\
\midrule
\midrule
Multiple Questions & \makecell[l]{\textbf{CHQ} \\\\\\\\ \textbf{FAQ}\\ \textbf{Ground truth} \\ \textbf{Prediction}} & \makecell[l]{Bipolar and Generalized Anxiety Disorder I read about Transcranial magnetic stimulation Therapy.\\ Do you know anything about it? Has it had success? Also wondering about ECT? ...\\ Is that true for mixed bipolar and generalized anxiety disorder along with meds?\\ Have you ever heard of this? \\ How effective is Transcranial magnetic stimulation for GAD? \\ True \\ False }  \\
\midrule
\midrule
Co-reference & \makecell[l]{\textbf{CHQ} \\ \\    \textbf{FAQ}\\ \textbf{Ground truth} \\ \textbf{Prediction}} & \makecell[l]{spina bifida; vertbral fusion;syrinx tethered cord.\\ can u help for treatment of these problem. \\ Does Spina Bifida cause vertebral fusion? \\ True \\ True }  \\ 
\midrule
\midrule
\end{tabular}
}
\caption{Qualitative analysis of the outputs produced by our RQE model. We categorize the errors into different buckets and provide cherry-picked examples to prove our claim. }
\label{table:qualitativerqe}
\end{center}
\end{table*}
\subsubsection{Results and Discussion}

The results over the validation set are in Table \ref{tab:rqevalid}. We see that the MT-DNN model performs the best amongst all the other models. Addition of the \textit{QQP} datasets did not add extra value. We hypothesize that this is due to lack of in-domain medical data in the QQP dataset.

 The results of the MT-DNN model with the different training settings can be seen in Table \ref{tab:rqetest}.  The test set comprises of 230 question pairs. We observe that the \textit{DataAug} setting where the MT-DNN model is trained on  in-domain validation set augmented with UMLS features, performs the best amongst all the strategies. Similar to the validation set, in this setting we also modify the test set with the UMLS features by augmenting it using the procedure of data augmentation described above. Therefore, the test set now comprises of 460 question pairs. We refer to the provided test set of 230 pairs as \textit{original} and the augmented test set as \textit{UMLS}. We submitted the outputs on both the original test set and the UMLS augmented test set and observe that the latter gives \textbf{+4.3} F1 gain over the original test set. We hypothesize that the addition of the UMLS augmented data in the training process helped the model to disambiguate false negatives.  

Despite training data being about medical questions, it has a different data distribution and language structure. Adding it actually harms the model, as seen by the \textit{ + Orig + DataAug + QQP} model. For our final submission, we took an ensemble of all submissions using a majority vote strategy. The ensemble model gave us the best performance.
\begin{table}
\small
\begin{center}{
  \begin{tabular}{|l|l|l|}
  \textbf{Model} &\textbf{Accuracy}  & \textbf{F1} \\
   \toprule
    \citet{abacha2016recognizing} & - & 75.0\\
   SVM & 71.9 & 70.0 \\
 \midrule
 BERT & 76.2 & 76.2  \\
 MT-DNN  + Orig  & 78.1 & \textbf{77.4} \\
 MT-DNN  + QQP  & \textbf{80.8}& 77.2 \\
  \bottomrule
  \end{tabular}
  }
  \caption{Results on the RQE validation set.}
  \label{tab:rqevalid}
  \end{center}
\end{table}

 	\begin{table}
 	\small
	\begin{center}{
	  \begin{tabular}{|l|l|l|}
	  & \textbf{Model} & \textbf{F1} \\
	   \toprule
	    & \citet{aclmediqa}  & 54.1 \\
	    \midrule
	 MT-DNN  & + Orig & 58.9\\
	 & + Orig + DataAug + QQP & 60.6\\
	  & + DataAug (UMLS) & \textbf{64.9} \\
	   & + DataAug (original) & 61.5 \\
	   & + DataAug + QQP (UMLS) & \textbf{64.9} \\
	 
	  & Ensemble & \textbf{65.8} \\
	  \bottomrule
	  \end{tabular}
	  }
	  \caption{Results on the RQE test set.}
	  \label{tab:rqetest}
	  \end{center}
	\end{table}

\subsubsection{Error Analysis}
\begin{table*}[t!]
\centering
\small
\begin{center}{
  \begin{tabular}{|c|c|c|c|}
    &  \textbf{  Questions }  &\textbf{Avg answer count } & \textbf{  Avg answer length } \\
   \toprule
   Train set 1  &  104 & 8 & 434.8\\ 
   Train set 2  &  104 & 8 & 432.5\\ 
   Validation set   &  25 & 9 & 420.4\\ 
   Test set  &  150 & 7 & 418.0\\ 
   \bottomrule
  \end{tabular}
  }
  \caption{Dataset statistics for re-ranking task.}
  \label{tab:rerankqastats}
  \end{center}
\end{table*}
Since we used the validation set for training the model, we cannot directly perform a standard error analysis. However, we manually analyze 100 question pairs from the test set and look at the different model predictions.  We categorize errors into the following categories, as shown in Table \ref{table:qualitativerqe}.
\begin{enumerate}
    \item \textbf{Lexical Overlap:} Most of the models we used above rely strongly on lexical overlap of tokens. Therefore, question-pairs with high orthography overlap have a strong prior for the \textit{True} label denoting entailment.
    \item \textbf{Multiple-Questions:}
    Often CHQ questions contained multiple sub-questions. We hypothesize that multiple questions tend to confuse the model. Furthermore, as seen in Table \ref{table:qualitativerqe}, the FAQ entails from two sub-questions in the CHQ. This shows that the model lacks the ability to perform multi-hop reasoning.
    \item \textbf{Co-reference:} The model is required to perform entity co-reference as part of the entailment. In the example shown in Table \ref{table:qualitativerqe}, majority of our models marked this as entailment purely because of lexical overlap. However, there was a need for the model to identify co-reference between \textit{these problem} and the problems mentioned in the previous sentence.
\end{enumerate}
\subsection{Question-Answering}
\label{qa}

In this section, we focus on building a re-ranker for question-answering systems. In particular, we attempt to use the NLI and RQE models for this task. In the ACL MediQA challenge, the question-answering system CHiQA \footnote{https://chiqa.nlm.nih.gov/} provides a possible set of answers and the task is to rank them in the order of relevance. 

\paragraph{Data:} The task-3 dataset comprises of 2 training sets and a validation set. The distribution of the data across train, validation and test was consistent in terms of average number of answer candidates and average answer length per questio can be seen in Table \ref{tab:rerankqastats}.

\subsubsection{Our Method} We implement the following re-ranking methods.
\paragraph{BM25:} This is a ranking algorithm used for relevance based ranking given query. The formulation is given below:

\begin{equation}\scriptsize
\operatorname{score}(D,Q)=\sum_{i=1}^{n} \operatorname{IDF}\left(q_{i}\right) \cdot \frac{f\left(q_{i}, D\right) \cdot\left(k_{1}+1\right)}{f\left(q_{i}, D\right)+k_{1} \cdot\left(1-b+b \cdot \frac{|D|}{\operatorname{avgd}}\right)}    
\end{equation}

\begin{equation}\small
    \operatorname{IDF}\left(q_{i}\right)=\log \frac{N-n\left(q_{i}\right)+0.5}{n\left(q_{i}\right)+0.5}
\end{equation}
Here \emph{D} is the answer. Q is a list of all words in the question. $q_{i}$ refers to a single word. $f(q_{i},D)$ is the term frequency of $q_{i}$ in document D. $\operatorname{avgd}$ is the average answer length.
The hyper-parameters used for this experiment were b = 0.75 and k1= 1.2. As shown in table \ref{tab:rereankqaacc} this gave an accuracy of 66.6 on the validation set.

\paragraph{NLI-RQE based model:} In our second approach we leverage the pre-built NLI and RQE models from Task 1 and 2 by including the NLI and RQE scores for each question-answer pair as a feature. For instance, given a question, for each answer snippet we compute NLI scores for each sentence in the answer with the question. Since the answer snippet also contains sub-questions, we use the RQE scores to compute entailment with the question. This is illustrated below:\\
\textbf{Question: }"about uveitis. IS THE UVEITIS, AN AUTOIMMUNE DISEASE" 

For the NLI scoring we would consider statements from the answer which might predict entail, contradict or neutral for the pair. Such as \textit{Uveitis is caused by inflammatory responses inside the eye.}

Similarly we use the question phrases from the answer to give the particular answer a RQE score based on the number of entailments \textit{Facts About Uveitis (What Causes Uveitis?)}

Finally, we use the BM25 score for the given answer and concatenate with the above features and use SVM as the classifier.


\begin{table}[h]
\small
\centering
 \begin{center}\resizebox{0.5\textwidth}{!}{
  \begin{tabular}{|c|c|}
 
   \textbf{Model} &  \textbf{Accuracy \%}  \\
   \toprule
  BM-25 & 66.6 \\
  RQE+NLI+Source & \textbf{67.5} \\
  \midrule
  \citet{aclmediqa} (Unseen Test) & 51.7 \\
  Ours & 56.5 \\
 \bottomrule
  \end{tabular}
  }
  \caption{Accuracy for task 3 on both validation set (top) and test set (bottom).}
  \label{tab:rereankqaacc}
  \end{center}
\end{table}
\subsubsection{Results}
Table \ref{tab:rereankqaacc} documents the results of our experiments. We observe that adding NLI and RQE as features 
show some improvement over the BM25 model. 

%% file: error-analysis.tex
\begin{table*}[htb]
\small
\begin{center}\resizebox{\textwidth}{!}{
\begin{tabular}{|l|l|l|} 

Lexical Overlap & \makecell[l]{\textbf{Premise} \\   \textbf{Hypothesis}\\ \textbf{Ground truth} \\ \textbf{Prediction}} & \makecell[l]{She is on a low fat diet \\ She said they also have her on a low salt diet. \\ Neutral \\ Entailment }  \\
\midrule
\midrule
Disease-Symptom relation & \makecell[l]{\textbf{Premise} \\   \textbf{Hypothesis}\\\\ \textbf{Ground truth} \\ \textbf{Prediction}} & \makecell[l]{Patient has diabetes \\ The patient presented with a change in mental status\\ and hyperglycemia. \\ Entailment \\ Neutral }  \\
\midrule
\midrule
Drug-Disease relation & \makecell[l]{\textbf{Premise} \\ \\    \textbf{Hypothesis}\\ \textbf{Ground truth} \\ \textbf{Prediction}} & \makecell[l]{She was treated with Magnesium Sulfate, Labetalol, Hydralazine\\ and bedrest as well as betamethasone. \\ The patient is pregnant \\ Entailment \\ Neutral }  \\ 
\midrule
\midrule
\end{tabular}
}
\caption{Qualitative analysis of the outputs produced by our model. We categorize the errors into different buckets and provide cherry-picked examples to demonstrate each category. }
\label{table:qualitative}
\end{center}
\end{table*}

%% file: Conclusion.tex
\section{Conclusion and Future Work}
In this work,  we present a multi-task learning approach for textual inference and question entailment tailored for the medical domain. We observe that incorporating domain knowledge for specialized domains such as the medical domain is necessary. This is because models such as BERT and MT-DNN have been pre-trained on large amounts of generic domains, leading to possible domain mismatch.  In order to achieve domain adaptation, we explore techniques such as data augmentation using UMLS features, abbreviation expansion and observe a gain of +10.8 F1 for RQE. There are still many standing challenges such as incorporating common-sense knowledge apart from domain knowledge and multi-hop reasoning which pose an interesting future direction. 

In the future, we also plan to explore other ranking methods based on relevancy feedback or priority ranking for task 3. We believe using MedQuad \cite{019arXiv190108079A} as training set could further help improve the performance.

\section*{Acknowledgement}
We are thankful to the anonymous reviewers for their valuable suggestions. 